\documentclass[letterpaper]{article} 
\usepackage{aaai2026}  
\usepackage{times}  
\usepackage{helvet}  
\usepackage{courier}  
\usepackage[hyphens]{url}  
\usepackage{graphicx} 
\usepackage{natbib}  
\usepackage{caption} 
\usepackage{algorithm}
\usepackage{algorithmic}

\usepackage{newfloat}
\usepackage{listings}
\DeclareCaptionStyle{ruled}{labelfont=normalfont,labelsep=colon,strut=off} 
\floatstyle{ruled}
\newfloat{listing}{tb}{lst}{}
\floatname{listing}{Listing}
\title{TimeBill: Time-Budgeted Inference for Large Language Models}
\author {
    Qi Fan, An Zou, Yehan Ma\thanks{Corresponding author} \\
}
\affiliations {
    Shanghai Jiao Tong University\\
    \{fanqi666, an.zou, yehanma\}@sjtu.edu.cn\\
}

\usepackage{amsmath}
\usepackage{amssymb}
\usepackage{booktabs}
\usepackage{multirow}
\usepackage{amsthm}
\usepackage{tikz}

\newcommand*\circled[1]{\tikz[baseline=(char.base)]{
            \node[shape=circle,draw,inner sep=0.2pt] (char) {#1};}}

\begin{document}

\maketitle

\begin{abstract}
    Large Language Models (LLMs) are increasingly deployed in time-critical systems, such as robotics, autonomous driving, embodied intelligence, and industrial automation, where generating accurate responses within a given time budget is crucial for decision-making, control, or safety-critical tasks. 
    However, the auto-regressive generation process of LLMs makes it challenging to model and estimate the end-to-end execution time. 
    Furthermore, existing efficient inference methods based on a fixed key-value (KV) cache eviction ratio struggle to adapt to varying tasks with diverse time budgets, where an improper eviction ratio may lead to incomplete inference or a drop in response performance. 
    In this paper, we propose TimeBill, a novel time-budgeted inference framework for LLMs that balances the inference efficiency and response performance. 
    To be more specific, we propose a fine-grained response length predictor (RLP) and an 
    execution time estimator (ETE) to accurately predict the end-to-end execution time of LLMs. 
    Following this, we develop a time-budgeted efficient inference approach that adaptively adjusts the KV cache eviction ratio based on execution time prediction and the given time budget. 
    Finally, through extensive experiments, we demonstrate the advantages of TimeBill in improving task completion rate and maintaining response performance under various overrun strategies.
\end{abstract}

\section{Introduction}

With the development of Large Language Models (LLMs), LLMs have been widely applied in time-critical systems, such as robotics, autonomous driving, embodied intelligence, and industrial automation. 
For instance, Autoware.Flex~\cite{Autoware.Flex} utilizes LLMs to translate natural language instructions into a format that autonomous driving systems can understand during the driving process. 
DriveGPT4~\cite{DriveGPT4} employs LLMs to perceive the driving environment and generate driving decisions along with corresponding explanations. 
In these scenarios, hard or firm deadlines may be introduced, especially when LLMs are involved in decision-making, control, or safety-critical tasks.
The inference of LLMs should be completed within a specific time budget while ensuring the performance of the response.
Therefore, it is essential to model the end-to-end execution time of LLM inference and balance between inference efficiency and response performance under the given time budget.
However, applying LLMs in hard real-time systems faces numerous challenges. 

Unlike Convolutional Neural Networks (CNNs), LLMs exhibit significant uncertainty in the end-to-end execution time due to the auto-regressive generation process~\cite{AR}.
Since the end-to-end execution time of LLMs is closely related to the number of generated tokens, namely response length, accurate modeling requires fine-grained prediction of response length. 
Several predictors rely on coarse-grained classification~\cite{ProxyModel-5class, S3-10class}, further increasing the error in modeling.
Moreover, the response length is influenced by a series of factors such as input content and the LLM itself~\cite{Inter-Intra-Model-Response-Difference}. 
Different input content can lead to various response lengths, while even with the same input, different LLMs may generate responses with diverse lengths. 
Therefore, a fine-grained predictor well-aligned with the target LLM to be deployed is crucial for accurately modeling the end-to-end execution time of LLMs.

\begin{figure}[t]
    \centering
    \includegraphics[width=0.8\linewidth]{figures/inference_comparison.pdf}
    \caption{An example of different inference strategies. The vanilla inference may overrun and miss the deadline, resulting in incomplete output. The As-Fast-As-Possible (AFAP) strategy will degrade the response performance, while time-budgeted inference improves the response performance under timing constraints.}
    \label{fig:inference_comparison}
\end{figure}

Existing efficient inference methods for LLMs can be categorized into two types: offline and online. 
Offline efficient inference methods, including quantization~\cite{Smoothquant, AWQ, GPTQ} and pruning~\cite{SparseGPT, LLM-Pruner}, compress the model before deployment to reduce resource consumption. 
However, they cannot adjust according to the time budget at runtime.
Online methods are orthogonal to the offline ones, involving key-value (KV) cache eviction~\cite{StreamingLLM, SnapKV, DuoAttention} and quantization~\cite{KVQuant, KIVI}. 
Furthermore, the impact of KV cache eviction varies across different tasks~\cite{SnapKV}, and different tasks may also have different time budgets. 
A fixed KV cache eviction ratio lacks the flexibility to handle such diverse scenarios. 
For instance, a higher ratio helps meet time budgets but harms response performance, while a lower ratio results in overruns.
Therefore, it is vital to develop an efficient inference method that dynamically adjusts the KV cache eviction ratio based on the time budget, enabling LLMs to complete inference on time while maintaining response performance.

In this paper, we propose TimeBill, a time-budgeted inference framework for LLMs. 
As shown in Fig.~\ref{fig:inference_comparison}, different from the As-Fast-As-Possible (AFAP) strategy, TimeBill balances inference efficiency and response performance. 
Beginning with presenting the problem formulation of time-budgeted inference for LLMs, we introduce a fine-grained response length predictor (RLP) and a workload-guided execution time estimator (ETE) to accurately predict the end-to-end execution time of LLMs. 
Furthermore, we develop a time-budgeted efficient inference method, which adapts the KV cache eviction ratio based on the execution time prediction and the time budget. 
Finally, we conduct a series of experiments and demonstrate the effectiveness of the TimeBill framework. 
The contributions are fourfold.

\begin{itemize}
    \item We present the problem formulation of time-budgeted inference for LLMs and introduce a novel framework named TimeBill, which balances the timing performance of inference and response performance.
    \item We construct a fine-grained response length predictor (RLP), providing precise response length prediction of the target LLM to be deployed.
    \item We propose a workload-guided execution time estimator (ETE) with analytical modeling and profiling integrated, offering accurate end-to-end execution time estimation.
    \item We develop a time-budgeted efficient inference mechanism, which effectively adjusts the KV cache eviction ratio based on the execution time prediction and time budget, thereby improving task completion rate and maintaining response performance.
\end{itemize}

\section{Related Work}
\label{sec:related}

\subsection{Execution Time Estimation}

Recently, real-time inference has been studied in deterministic DNNs, ensuring strict time constraints in time-critical applications. 
Chen et al.~\cite{SCENIC} map the fully-connected DNNs to a segmented task model on heterogeneous platforms.
Kang et al.~\cite{Lalarand} establish response time analyses of layered DNNs and introduce quantization and runtime layer migration to reduce inference time.
However, unlike the fixed-structure deterministic DNN, the end-to-end execution time of LLMs exhibits uncertainty due to the auto-regressive generation~\cite{AR}, which makes the execution time dependent on the dynamic response lengths. 

A series of response length predictors have been proposed.
For instance, PiA~\cite{PiA-self-regression} uses fine-tuning or prompt engineering to enable the target LLM to predict its response length before answering the question. 
ProxyModel~\cite{ProxyModel-5class} builds a 5-class classifier based on BERT~\cite{BERT} to predict which bucket the response length will fall into. 
S$^{3}$~\cite{S3-10class} uses DistilBERT~\cite{DistilBERT} to construct a 10-class classifier. 
However, BERT-based predictors struggle with long input content. 
On the other hand, coarse-grained predictors cannot provide precise predictions for response time estimation.
Therefore, a fine-grained response length predictor that can handle long input sequences is crucial to perform accurate response time prediction.

In addition, some works explore predicting execution time based on the execution characteristics of LLMs. 
For example, RLM-ML~\cite{RLM-predicting} and LLMStation~\cite{LLMStation} combine the roofline model with machine learning to predict execution time through data collection and optimization. 
BestServe~\cite{Bestserve} uses an adaptive roofline model for prediction. 
However, ML-based execution time prediction methods lack interpretability and are not friendly to online prediction.

\subsection{Efficient LLM Inference}

Due to the high computational resource usage and inference latency, LLMs face challenges in real-time applications. 
To this end, a series of efficient inference methods for LLMs have been proposed, which are mainly divided into two categories: offline and online methods.

Offline methods compress the model before deployment for lower resource consumption and time cost during inference. 
For example, SmoothQuant~\cite{Smoothquant} quantizes both weights and activations, while AWQ~\cite{AWQ} and GPTQ~\cite{GPTQ} perform weight-only quantization. 
In addition, SparseGPT~\cite{SparseGPT} and LLM-Pruner~\cite{LLM-Pruner} apply pruning to the model weights. 
However, when facing varying time costs due to the dynamic response time during runtime, these methods are unable to adjust according to the time budgets.

Online methods achieve efficient inference primarily through runtime eviction and quantization of the key-value (KV) cache. 
For instance, StreamingLLM~\cite{StreamingLLM}, SnapKV~\cite{SnapKV}, and DuoAttention~\cite{DuoAttention} discard less important parts of the KV cache. 
Meanwhile, KVQuant~\cite{KVQuant} and KIVI~\cite{KIVI} quantize the KV cache to 4 bits or lower. 
Although online methods allow runtime adjustments, they overlook the time budgets, which may lead to overruns or inaccurate responses. 
Therefore, an efficient inference method that accounts for the time budget while maintaining the response performance is essential.

\section{Overview}
\label{sec:overview}

In this section, we present the problem formulation of time-budgeted inference for LLMs and our TimeBill framework.

\subsection{Time-Budgeted Inference Problem for LLMs}
\label{sec:problem_formulation}

The inference process of LLMs consists of two phases, namely the prefill phase and the decoding phase~\cite{DistServe}. 
During the prefill phase, LLMs process the input prompt $x$ and produce the first output token $\hat{y}_0$. 
LLMs perform autoregressive generation in the subsequent decoding phase, which comprises $N-1$ sequential decoding steps, as shown in Fig.~\ref{fig:timebill_framework}. 
In each decoding step, LLMs generate a new output token $\hat{y}_i$ based on the previously generated tokens, where $i\in[1, N-1]$ is the index of the decoding step.

In time-critical systems with hard deadline constraints (hard real-time systems), inference that exceeds the time budget $T$ is considered a system failure~\cite{Overrun}. 
Therefore, the goal of time-budgeted inference for LLMs is to optimize the response performance while ensuring that inference completes within the time budget. 
To be more specific, the time-budgeted inference for LLMs can be formulated as
\begin{small}
\begin{subequations}
    \begin{align}
        \underset{\theta}{\max} \quad &\mathcal{M}(\hat{\mathbf{y}}(\theta), \mathbf{y}) \label{eq:problem_object} \\
        \text{s.t. } \quad & t_{\text{e2e}}(x, \theta) \leqslant T \label{eq:problem_time_constraint} \\
        & N \leqslant N_{\text{max}}.\label{eq:problem_length_constraint}\end{align}\label{eq:problem}\end{subequations}\end{small}
Given LLMs, the execution time and response performance of generating $\hat{y}_i$ are affected by a series of configuration factors $\theta$ in the decoding phase, such as the KV cache eviction ratio $\alpha$. 
The objective in \eqref{eq:problem_object} is to configure $\theta$ at run-time for each LLM inference in order to optimize the 
response performance metric  $\mathcal{M}(\cdot)$ of generated response content $\hat{\mathbf{y}}(\theta) = (\hat{y}_0, \hat{y}_1, \ldots, \hat{y}_i, \ldots, \hat{y}_{N-1})$ compared with the ground truth $\mathbf{y} = ({y}_0, \ldots, {y}_{N-1})$.
And the end-to-end execution time is $t_{\text{e2e}}$ that should stay within the given time budget $T$, which is described as constraint \eqref{eq:problem_time_constraint}. 
The inference process completes when a termination token is generated or the maximum generation length $N_{\text{max}}$ is reached, i.e., $N \leqslant N_{\text{max}}$ in \eqref{eq:problem_length_constraint}. 

\subsection{TimeBill Framework}
\label{sec:framework}

\begin{figure}[t]
    \centering
    \includegraphics[width=0.95\linewidth]{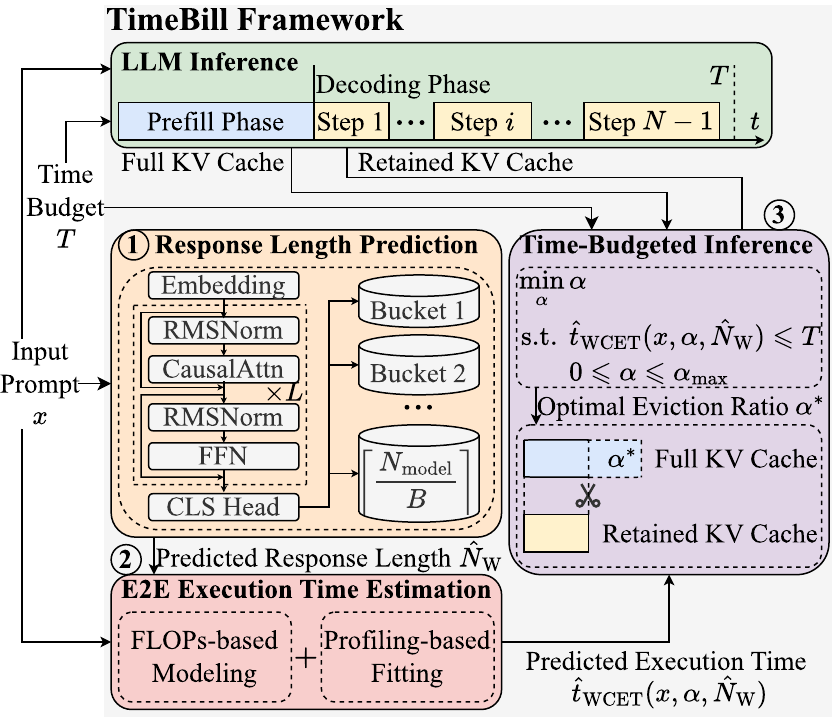}
    \caption{The overview of the TimeBill framework.}
    \label{fig:timebill_framework}
\end{figure}

Challenges of the time-budgeted inference problem for LLMs described in Prob.~\eqref{eq:problem} are:

\noindent \textit{\textbf{Challenge 1 {[Run-time execution time $t_{\text{e2e}}$ Estimation]}}: 
$t_{\text{e2e}}$ largely depends on the response length $N$, which cannot be established until the inference is done. Therefore, the run-time response length prediction is the first challenge. In addition, given the response length prediction, how to accurately estimate $t_{\text{e2e}}$ is another challenge.}

\noindent \textit{\textbf{Challenge 2 {[Run-time LLM configuration $\theta$]}}: How to map the execution time estimation and decoding phase configuration $\theta$ is the first challenge. Furthermore, how to configure the $\theta$ at runtime to optimize the response performance $\mathcal{M}(\cdot)$ within a certain time budget $T$ is another challenge.}

We propose the TimeBill framework to address the above challenges. 
The overview of the TimeBill framework is presented in Fig.~\ref{fig:timebill_framework}. 
\circled{1} The fine-grained RLP based on Small Language Model (SLM) is proposed to predict response lengths of the target LLM inference, which will be introduced in Sec.~\ref{sec:predictor}.
\circled{2} Given the predicted response length, the workload-guided ETE is constructed by integrating FLOPs analysis and profiling to offer accurate end-to-end execution time estimation, 
which will be introduced in Sec.~\ref{sec:execution}.
\circled{3} Given the time budget and estimated execution time for each LLM inference,
we develop a time-budgeted efficient inference mechanism based on the KV cache eviction. 
The mechanism establishes the optimal KV cache eviction ratio to maximize the response performance $\mathcal{M}(\cdot)$ within the time budget, which will be described in Sec.~\ref{sec:inference}.

\section{Fine-grained Execution Time Prediction for LLM}
\label{sec:time_prediction}

According to Challenge 1 mentioned in Sec.~\ref{sec:framework}, the accurate end-to-end execution time prediction is the prerequisite for time-budgeted LLM inference. 
This section introduces the fine-grained RLP to predict the response length $N$, and the workload-guided  ETE to estimate the end-to-end execution time $t_{\text{e2e}}$. 

\subsection{Fine-grained Response Length Predictor}
\label{sec:predictor}

\subsubsection{Predictor Design}

\begin{figure}[t]
    \centering
    \includegraphics[width=0.86\linewidth]{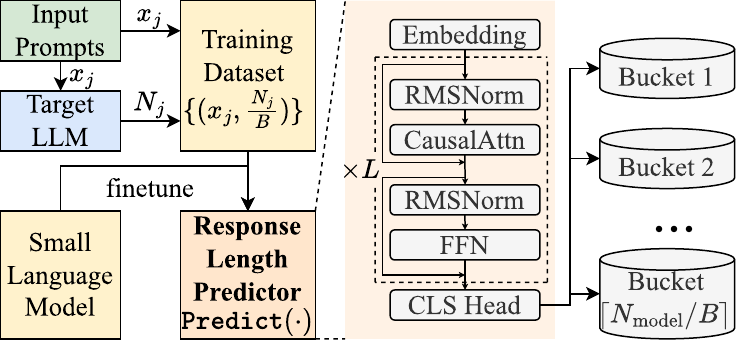}
    \caption{The overview of the proposed fine-grained response length predictor (RLP).}
    \label{fig:predictor}
\end{figure}

To provide accurate end-to-end execution time estimation, we develop a fine-grained RLP. 
We define the response length prediction as a classification task instead of a regression task, since predicting the exact response length is challenging.
Hence, the RLP needs to determine which bucket the response length will fall into, where the bucket size is fixed at $B$, as shown in Fig.~\ref{fig:predictor}.
Due to the limited context length of BERT~\cite{BERT}, it is difficult to process longer inputs.
Therefore, the RLP $\texttt{Predict}(\cdot)$ is based on a Small Language Model (SLM) to better process long input prompts, which has significantly fewer parameters than the target LLM. 
The architecture of RLP is shown in Fig.~\ref{fig:predictor}, which consists of an embedding layer, $L$ decoder layers, and a classification head. 
Each decoder layer includes four layers in sequence: an \texttt{RMSNorm} layer~\cite{RMSNorm}, a \texttt{CausalAttention} layer, another \texttt{RMSNorm} layer, and an \texttt{FFN with SwiGLU} layer~\cite{SwiGLU}.

Since the response length largely depends on the target LLM~\cite{Inter-Intra-Model-Response-Difference}, we employ the knowledge distillation approach to better align the RLP with the target LLM. 
As shown in Fig.~\ref{fig:predictor}, for a given input prompt $x_j$, we collect the actual length $N_j$ of the response generated by the target LLM, where $j$ is the index of the data item.
Therefore, we construct the training dataset consisting of pairs $(x_j, \lceil \frac{N_j}{B} \rceil)$, where $\lceil \frac{N_j}{B} \rceil$ represents the target bucket index, namely the classification label.  
According to the model card, the maximum generation capacity of the target LLM is $N_{\text{model}}$ tokens, which exceeds the maximum generation length $N_{\text{max}}$ specified during runtime, namely $N_{\text{model}} \geqslant N_{\text{max}}$.
Therefore, there are $\lceil N_{\text{model}} / B \rceil$ buckets for classification during the training process. 

According to the constraint~\eqref{eq:problem_length_constraint}, we perform post-processing to limit the maximum predicted response length by $N_{\text{max}}$. 
Assuming the predicted bucket index is $\hat{n} = \texttt{Predict}(x)$, indicating the response length is between $(\hat{n}-1)B$ and $\hat{n}B$, and $\hat{n}B$ is reported as the predicted response length. The predicted response length $\hat{N}$ after post-processing can be obtained as follows: 
\begin{small}
\begin{equation}
    \label{eq:predictor_inference}
    \hat{N} = \min(N_{\text{max}}, \hat{n}B)=\min(N_{\text{max}}, \texttt{Predict}(x)B)
\end{equation}
\end{small}

\subsection{Workload-guided Execution Time Estimator}
\label{sec:execution}

End-to-end execution time estimation is essential in hard real-time systems~\cite{Overrun}. 
Modeling the worst-case execution time (WCET) during system design ensures that each LLM inference meets deadlines. 
In this section, we develop a workload-guided ETE, integrating floating point operations (FLOPs) -based analytical modeling and profiling-based fitting.

\subsubsection{FLOPs-based Modeling}

We adopt FLOPs-based modeling to analyze the relationship between computational workload
and WCET, providing theoretical support for profiling-based fitting. 
Since most modern Transformer-based LLMs~\cite{LLaMA3, Qwen2.5} employ an architecture comprising an embedding layer, a series of decoder layers, and a language modeling head. 
The architecture of each decoder layer is \texttt{Norm\allowbreak-CausalAttention\allowbreak-Norm\allowbreak-FeedForward}.
Given that matrix multiplications account for the majority of FLOPs~\cite{MegatronLM}, since there is no matrix multiplication in the embedding and \texttt{Norm} layer, we only analyze the FLOPs from the \texttt{CausalAttention} and 
\texttt{FeedForward} layers and the language modeling head \texttt{LMHead}.
We suppose the number of FLOPs of each layer in the prefill phase and decoding step is denoted as $f_{\text{prefill-phase}}^{\texttt{LayerName}}$ and $f_{\text{decoding-step}}^{\texttt{LayerName}}$, respectively.

The main computation of the \texttt{CausalAttention} layer is $\text{CausalAttention}(Q, K, V) = \text{softmax}\left(\frac{QK^\top \odot M}{\sqrt{d}}\right)V$, where $d$ is the hidden size, $\odot$ denotes element-wise multiplication, and $M$ is the causal attention mask~\cite{Attention}. 
In the prefill phase, both $Q$ and $K$ have length $N_x$, where $N_x$ is the length of the input prompt $x$. 
Therefore, $f_{\text{prefill-phase}}^{\texttt{CausalAttention}}$ is \emph{\textbf{quadratic}} in $N_x$. 
In the decoding step, only the last generated token needs processing, so the length of $Q$ is 1 and the length of $K$ is $N_{\text{kv}}$, which denotes the current length of the KV cache. 
As a result, $f_{\text{decoding-step}}^{\texttt{CausalAttention}}$ is \emph{\textbf{linear}} in $N_{\text{kv}}$. 
Similarly, the \texttt{FeedForward} layer handles the input prompt $x$ consisting of $N_x$ tokens during the prefill phase, and processes the last generated token in each decoding step.
Since the \texttt{FeedForward} layer processes each input token independently~\cite{Orca}, $f_{\text{prefill-phase}}^{\texttt{FeedForward}}$ is \emph{\textbf{linear}} in $N_x$, while $f_{\text{decoding-step}}^{\texttt{FeedForward}}$ is \emph{\textbf{constant}} and determined solely by the hyperparameters of the model (e.g., hidden size, intermediate size).
The language modeling head \texttt{LMHead} takes the hidden state of the last input token as the input and produces the logits of the next token. 
Therefore, the FLOPs of the \texttt{LMHead} layer are solely related to the model architecture, regardless of the inference stage, namely $f_{\text{prefill-phase}}^{\texttt{LMHead}}$ and $f_{\text{decoding-step}}^{\texttt{LMHead}}$ are \emph{\textbf{constant}}.

Therefore, the number of FLOPs in the prefill phase $f_{\text{prefill-phase}}$ is quadratic in $N_x$ with linear and constant terms included, while that in the decoding step $f_{\text{decoding-step}}$ is linear in $N_{\text{kv}}$, namely
\begin{small}
\begin{equation}
    f_{\text{stage}} = f_{\text{stage}}^{\texttt{FeedForward}} + f_{\text{stage}}^{\texttt{CausalAttention}} + f_{\text{stage}}^{\texttt{LMHead}},  \\
\end{equation}\end{small} where stage is prefill phase or decoding step. 
Since the number of FLOPs and the corresponding execution time share the same form~\cite{PaLM}, we can derive the estimated execution time of the prefill phase $\hat{t}_{\text{prefill-phase}}$ with respect to $N_x$ (the length of input prompt $x$), and that of the $i$-th decoding step $\hat{t}_{\text{decoding-step}}^{i}$ with respect to $N_{\text{kv}}^{i}$ as follows:
\begin{small}\begin{subequations}\label{eq:exeuction_time_part}\begin{align}      &\hat{t}_{\text{prefill-phase}}(x) = a N_{x}^2 + b N_x + c \label{eq:prefill_phase_time_raw} \\
        &\hat{t}_{\text{decoding-step}}^{i}(N_{\text{kv}}^{i}) = pN_{\text{kv}}^{i} + q,\label{eq:decoding_step_time_raw}\end{align}\end{subequations}\end{small}
where $a$, $b$, $c$, $p$, and $q$ are corresponding coefficients. In the next subsection, we will present a profiling-based and data-driven approach to establish these coefficients. 
Furthermore, we observe in Eq.~\eqref{eq:decoding_step_time_raw} that $N_{\text{kv}}^{i}$ largely affects the execution time of each decoding step, the derivation and configuration of which will be discussed in detail.

\subsubsection{Profiling-based Fitting}

Although FLOPs-based analytical modeling characterizes computational workload in terms of $N_x$ and $N_{\text{kv}}^{i}$, actual execution time depends on implementation and hardware~\cite{PaLM}. 
To this end, we propose a profiling-based fitting method to estimate the execution time. To be more specific, we establish coefficients in Eq.~\eqref{eq:exeuction_time_part} using data-driven approaches based on execution time profiling, given certain LLM models and hardware platforms. 
Taking the prefill phase as an example,
the actual execution time $t_{\text{prefill-phase}}(N_x)$ for given $N_x$ can be measured. 
Hence, a dataset with pairs $(N_x, t_{\text{prefill-phase}}(N_x))$ can be obtained to derive the coefficients $a$, $b$, and $c$ in Eq.~\eqref{eq:prefill_phase_time_raw}.
Well-established fitting methods, such as the Least Squares (LS), can be applied. 
The same applies to the decoding step.

\subsubsection{End-to-end Execution Time Prediction}
According to Eq.~\eqref{eq:decoding_step_time_raw},  the execution time of each decoding step depends on $N_{\text{kv}}^{i}$.
Therefore, we explore the impact of the KV cache eviction ratio $\alpha$ on the execution time. 
Since KV cache eviction occurs after the prefill phase and before the decoding phase, a fraction $\alpha$ of the KV cache produced in the prefill phase will be evicted. 
As a result, in the $i$-th decoding step, the length of the KV cache is
\begin{small}\begin{equation}
\label{eq:N_kv}
N_{\text{kv}}^{i}(x, \alpha) = (1-\alpha)N_x + i-1. 
\end{equation}\end{small}
Given Eqs.~\eqref{eq:decoding_step_time_raw} and \eqref{eq:N_kv}, the estimated execution time of the $i$-th decoding step becomes
\begin{small}\begin{equation}
\label{eq:decoding-step}
\hat{t}_{\text{decoding-step}}^{i}(x, \alpha) = p((1-\alpha)N_x + i-1) + q.
\end{equation}\end{small}

According to the predicted response length $\hat{N}$, which is obtained according to Eq.~\eqref{eq:predictor_inference},
the estimated execution time of the decoding phase is the sum of the execution time of all decoding steps, i.e.,
\begin{small}\begin{equation}
\label{eq:decoding-phase}
 \hat{t}_{\text{decoding-phase}}(x, \alpha, \hat{N}) = \sum_{i=1}^{\hat{N}-1} \hat{t}_{\text{decoding-step}}^{i}(x, \alpha).   
\end{equation}\end{small}
The end-to-end execution time prediction consists of the estimated execution time of the prefill and decoding phases. 
Thus, we estimate the execution time based on Eqs.~\eqref{eq:prefill_phase_time_raw} and \eqref{eq:decoding-phase}, i.e., 
\begin{small}\begin{equation}
    \label{eq:execution_time}
    \begin{aligned}
        &\hat{t}_{\text{e2e}}(x, \alpha, \hat{N}) = \hat{t}_{\text{prefill-phase}}(x) + \hat{t}_{\text{decoding-phase}}(x, \alpha, \hat{N}). \\
    \end{aligned}
\end{equation}\end{small}

Furthermore, considering WCET, we introduce the pessimistic factor $k$, $k\geqslant 1$, to $\hat{N}$. Hence, the pessimistic predicted response length becomes $k\hat{N}$.
According to constraint \eqref{eq:problem_length_constraint}, the pessimistic predicted response length is $\hat{N}_{\text{W}} = \min(k\hat{N}, N_{\text{max}})$. 
Based on Eq.~\eqref{eq:execution_time}, we can estimate the WCET of executing a LLM inference as
\begin{small}\begin{equation}
    \label{eq:wcet}
    \hat{t}_{\text{WCET}}(x, \alpha, \hat{N}_{\text{W}}) = \hat{t}_{\text{prefill-phase}}(x) + \hat{t}_{\text{decoding-phase}}(x, \alpha, \hat{N}_{\text{W}}).
\end{equation}\end{small}
We will evaluate the impacts of $k$ in the evaluation.

\section{Time-Budgeted Efficient LLM Inference}
\label{sec:inference}

This section presents the mechanism and corresponding system deployment for time-budgeted LLM inference.

\subsection{Time-Budgeted LLM Inference Mechanism}

According to Challenge 2 mentioned in Sec.~\ref{sec:framework}, the configuration factors $\theta$ need to be adjusted at runtime due to the variances of input and response length. 
Therefore, we develop a time-budgeted efficient inference mechanism based on the KV cache eviction. 
Consequently, in the following analysis, we target the KV cache eviction ratio $\alpha$ as the configuration factor.

Since the response performance metric $\mathcal{M}(\cdot)$ is unknown until the inference is completed, which poses a challenge for maximizing the objective \eqref{eq:problem_object} in Prob.~\eqref{eq:problem}.
In general, the response performance is non-increasing as the KV cache eviction ratio $\alpha$ increases~\cite{DuoAttention}. 
Thus, the objective function can be transformed from maximizing the response performance to minimizing the KV cache eviction ratio $\alpha$.

Additionally, according to the constraint \eqref{eq:problem_time_constraint}, the end-to-end execution time $t_{\text{e2e}}$ should stay within the time budget $T$, which cannot be established until the inference is completed as well. 
To this end, we use the predicted worst-case execution time $\hat{t}_{\text{WCET}}$ in Eq.~\eqref{eq:wcet} as a conservative prediction of $t_{\text{e2e}}$. 
Additionally, the overall overhead of executing $\texttt{Predict}(\cdot)$ and $\hat{t}_{\text{WCET}}$ prediction cannot be ignored, which we denote as $t_{\texttt{Predict}}(x)$. 
Since $\hat{t}_{\text{WCET}}$ prediction is the prerequisite for establishing $\alpha$, the $t_{\texttt{Predict}}(x)$ can be measured directly.
Therefore, the constraint ~\eqref{eq:problem_time_constraint} is transformed to $t_{\texttt{Predict}}(x) + \hat{t}_{\text{WCET}}(x, \alpha, \hat{N}_{\text{W}}) \leqslant T$.

Therefore, we can obtain the converted problem of time-budgeted LLM inference, i.e., 
\begin{small}\begin{subequations}
    \label{eq:problem_converted}
    \begin{align}
        \underset{\alpha}{\min}\quad & \alpha \label{eq:problem_converted_object} \\
        \text{s.t. }\quad & t_{\texttt{Predict}}(x) + \hat{t}_{\text{WCET}}(x, \alpha, \hat{N}_{\text{W}}) \leqslant T \label{eq:problem_converted_time} \\
        & 0 \leqslant \alpha \leqslant \alpha_{\text{max}}. \label{eq:problem_converted_alpha}
    \end{align}
\end{subequations}\end{small}
Since an excessively large $\alpha$ may degrade $\mathcal{M}(\cdot)$ significantly, the maximum eviction ratio $\alpha_{\text{max}}$ is introduced in constraint \eqref{eq:problem_converted_alpha}. 
Substituting Eqs.~\eqref{eq:decoding-step}, \eqref{eq:decoding-phase}, and \eqref{eq:wcet} into Eq.~\eqref{eq:problem_converted_time}, we can derive the optimal $\alpha^*$ by solving Prob.~\eqref{eq:problem_converted}. 
\begin{small}\begin{equation}
    \label{eq:alpha_with_predictor}
    \begin{aligned}
        \alpha^* = \min \Bigg(\alpha_{\text{max}}, 1 &- \dfrac{T-\hat{t}_{\text{prefill-phase}}(x) - t_{\texttt{Predict}}(x)}{p N_x (\hat{N}_{\text{W}} - 1)}\\ &\quad\quad\quad\quad\quad\quad + \dfrac{\hat{N}_{\text{W}} - 2}{2pN_x} + \dfrac{q}{pN_x}\Bigg),
    \end{aligned}
\end{equation}\end{small}\noindent where $\hat{t}_{\text{prefill-phase}}(x)$ is defined in Eq.~\eqref{eq:prefill_phase_time_raw}. 
To summarize, given the input prompt $x$ and optional maximum generation length $N_{\text{max}}$, 
the coefficients in Eq.~\eqref{eq:exeuction_time_part}, and the pessimistic factor $k$, the optimal KV cache eviction ratio $\alpha^*$ for the time-budgeted efficient inference can be established to optimize the response performance within the time budget $T$. 

\subsection{System Deployment}
\label{sec:deployment}

\begin{figure}[b]
    \centering
    \includegraphics[width=0.86\linewidth]{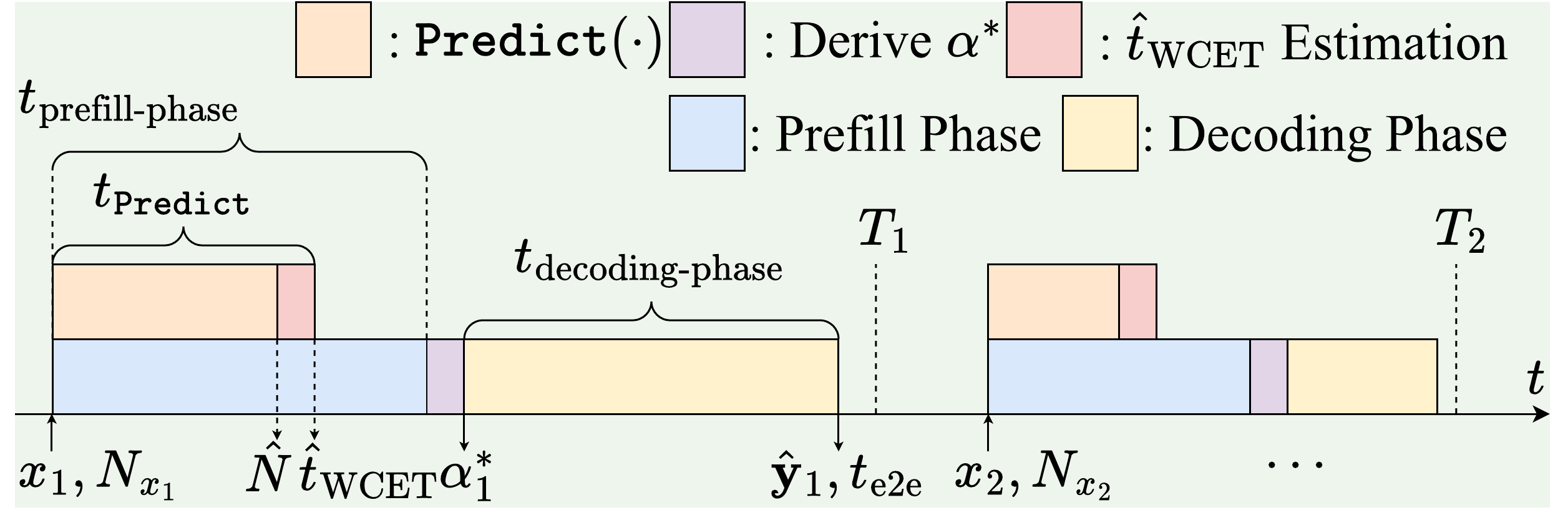}
    \caption{The timeline of TimeBill, where incoming arrows represent inputs (e.g., $x_1, N_{x_1}$) , and outgoing arrows represent outputs (e.g., $\hat{\mathbf{y}}_1, \alpha_1^*$).}
    \label{fig:deployment}
\end{figure}

As shown in Fig.~\ref{fig:deployment}, for each LLM inference, upon the input prompt $x$ arriving,  $N_x$ is known. 
Subsequently, an offline-trained RLP $\texttt{Predict}(\cdot)$ is utilized to predict $\hat{N}$ according to Eq.~\eqref{eq:predictor_inference}.
Based on $\hat{N}$,  $k$, and the offline-obtained coefficients in Eq.~\eqref{eq:exeuction_time_part}, $\hat{t}_{\text{WCET}}$ is estimated through Eq.~\eqref{eq:wcet}. 
At the same time, the LLM inference begins the prefill phase, which takes $t_{\text{prefill-phase}}$. 
After the prefill phase, 
$\alpha^*$ can be determined through Eq.~\eqref{eq:alpha_with_predictor}. $\alpha^*$ percent of the KV cache is evicted, and 
the retained KV caches are utilized during the subsequent decoding phase. 
After $t_{\text{decoding-phase}}$, the inference is completed, 
and the response $\hat{\mathbf{y}}$ is acquired. 

Since $\hat{t}_{\text{WCET}}$ is the prerequisite for establishing $\alpha^*$, $\hat{N}$ needs to be obtained before the decoding phase. 
Therefore, $\texttt{Predict}(\cdot)$ and $\hat{t}_{\text{WCET}}$ prediction can be executed in parallel with the prefill phase of the LLM on other processors, such as CPU or GPU.
If $ t_{\texttt{Predict}}\leqslant  t_{\text{prefill-phase}}$, the negative impact of $ t_{\texttt{Predict}}$ on response performance can be eliminated, i.e. $ t_{\texttt{Predict}}=0$ in Eq.~\eqref{eq:alpha_with_predictor}.
Therefore, in this work, we integrate our ETE in Sec.~\ref{sec:time_prediction} with prompt compression~\cite{PromptCompression}.
Note that the overhead of  Eq.~\eqref{eq:wcet} is ignorable compared with $\texttt{Predict}(\cdot)$.
Similar to 
Eq.~\eqref{eq:prefill_phase_time_raw},  the execution time of $\texttt{Predict}(\cdot)$ can be modeled as $\hat{t}_{\texttt{Predict}}(x_p) = a_p N_p^2 + b_p N_p + c_p$, where $x_p$ is the input of the predictor and $N_p$ is the length of $x_p$. 
Given 
$\hat{t}_{\text{prefill-phase}}$ obtained through Eq.~\eqref{eq:prefill_phase_time_raw}, an upper bound of $N_p$ can be derived by solving $\hat{t}_{\texttt{Predict}}(x_p) \leqslant \hat{t}_{\text{prefill-phase}}(x)$. 
Accordingly, prompt compression is performed to compress the input prompt $x$ into a shorter $x_p$.

Note that the time budget $T$ can vary across the inferences, e.g., $T_1\neq T_2$ in Fig.~\ref{fig:deployment}. 
Since $\hat{N}$, $\hat{t}_{\text{WCET}}$, and $\alpha^*$ are established for each inference, this can be handled naturally by TimeBill.

\section{Evaluation}
\label{sec:exper}
We first evaluate 
the efficacy of RLP in Sec.~\ref{sec:predictor_case}. 
And we demonstrate the performance of the ETE in Sec.~\ref{sec:time_case}.
Then, we compare TimeBill with several state-of-the-art approaches in Sec.~\ref{sec:compare_result}.
The impact of the pessimistic factor is discussed in Sec.~\ref{sec:exper_pess_factor}.

\subsection{Experimental Setup}

We utilize Qwen2.5-7B-Instruct~\cite{Qwen2.5} as the target LLM with a context length of 32,768 tokens and a maximum generation capacity $N_{\text{model}} = \text{8,192}$ tokens. 
The test dataset is LongBench~\cite{LongBench}, and the response performance score is evaluated using the official evaluation metrics, such as F1 score, ROUGE-L~\cite{ROUGE}, and Levenshtein distance. 
The KV cache eviction is performed using SnapKV~\cite{SnapKV}.
The experiments are conducted on a server equipped with Intel(R) Xeon(R) Platinum 8350C and an NVIDIA A40 GPU. 

\subsubsection{TimeBill Implementation}

We employ Qwen2.5-0.5B-Instruct~\cite{Qwen2.5} as the SLM to build the proposed RLP $\texttt{Predict}(\cdot)$ with default 512 buckets (the number of buckets will be discussed in Sec.~\ref{sec:predictor_case}). 
To avoid training on the test set, the prompts from the Arena-Human-Preference-100k dataset~\cite{predictor-dataset1, predictor-dataset2} are used to construct the dataset for training the $\texttt{Predict}(\cdot)$. 
The execution time of the target LLM is profiled to 
build the ETE, where $N_x$ and $N_{\text{kv}}$ range from 0 to 32,768.
The default pessimistic factor $k$ is set to 5 (the value of $k$ will be discussed in Sec.~\ref{sec:exper_pess_factor}). 
The maximum KV cache eviction ratio $\alpha_{\text{max}}$ is set to $95\%$.

\subsubsection{Approaches}

\begin{itemize}
    \item Vanilla inference (Vanilla), where the target LLM is directly used for inference.
    \item KV cache eviction with fixed $\alpha$~\cite{KNorm, H2O}, where $\alpha$ is set to $25\%, 50\%, 75\%, 95\%$, respectively. We denote them as $\alpha= \text{x} \%$ in this section.
     \item Activation-aware Weight Quantization (AWQ), where the model weights is quantized to 4 bits~\cite{AWQ}.
    \item Our proposed \textbf{TimeBill}.
\end{itemize}

\subsubsection{Overrun Strategies}

When an inference job is about to overrun and miss the deadline, the hard real-time system will apply an overrun strategy.  
We apply two of the most commonly used overrun strategies~\cite{Overrun}, 
\begin{itemize}
    \item \textit{Kill}. The current job will be terminated and considered incomplete. The response is regarded as empty since it misses the deadline. 
    \item \textit{Skip-Next}. Skip the next few jobs until the current job completes. The skipped jobs are considered incomplete and do not any produce response.
\end{itemize}

The average response performance score across all data items in the test set and the completion rate are reported, where the completion rate is defined as the percentile of the number of completed jobs to the total number of jobs.

\subsection{Efficacy of the Response Length Predictor}
\label{sec:predictor_case}

\begin{table}[b]
    \centering
    \small
    \setlength{\tabcolsep}{1mm}
    \begin{tabular}{@{}ccccccc@{}}
        \toprule
        Methods   & \multicolumn{4}{c}{\textbf{Ours}}       & ProxyModel & S3     \\ \midrule
        \#Buckets & Reg    & 128   & 256   & \textbf{512}   & 5          & 10     \\ \midrule
        MAE       & 64.21  & 48.95 & 44.15 & \textbf{42.71} & 105.72     & 108.96 \\ \midrule
        RMSE      & 103.30 & 87.57 & 78.63 & \textbf{78.13} & 136.79     & 148.91 \\ \midrule
        R$^2$     & 0.516  & 0.652 & 0.719 & \textbf{0.723} & 0.152      & -0.004 \\ \bottomrule
    \end{tabular}
    \caption{The efficacy of different response length predictors.}
    \label{tab:predictor_performance}
\end{table}

We compare our fine-grained RLP $\texttt{Predict}(\cdot)$ with BERT-based predictors, including ProxyModel~\cite{ProxyModel-5class} and S{$^3$}~\cite{S3-10class}. 
We test the $\texttt{Predict}(\cdot)$ with different bucket sizes ($B = 16, 32, 64$), which correspond to 512, 256, and 128 buckets, respectively.
In addition, we test $\texttt{Predict}(\cdot)$ modeled in a regression manner.
We evaluate the prediction error using metrics including Mean Absolute Error (MAE), Root Mean Square Error (RMSE), and R-squared (R{$^2$}), where the ground-truth response lengths are collected using the target LLM. 
As shown in Tab.~\ref{tab:predictor_performance}, $\texttt{Predict}(\cdot)$ outperforms ProxyModel and S3.
The $\texttt{Predict}(\cdot)$ based on regression modeling performs poorly, indicating predicting exact response length directly is challenging. 
$\texttt{Predict}(\cdot)$ with 512 buckets achieves the best performance. 
As the number of buckets increases, the granularity of $\texttt{Predict}(\cdot)$ becomes finer, and the performance improves. 

\subsection{Performance of the Execution Time Estimator}
\label{sec:time_case}

\begin{figure}[t]
    \centering
    \includegraphics[width=0.88\linewidth]{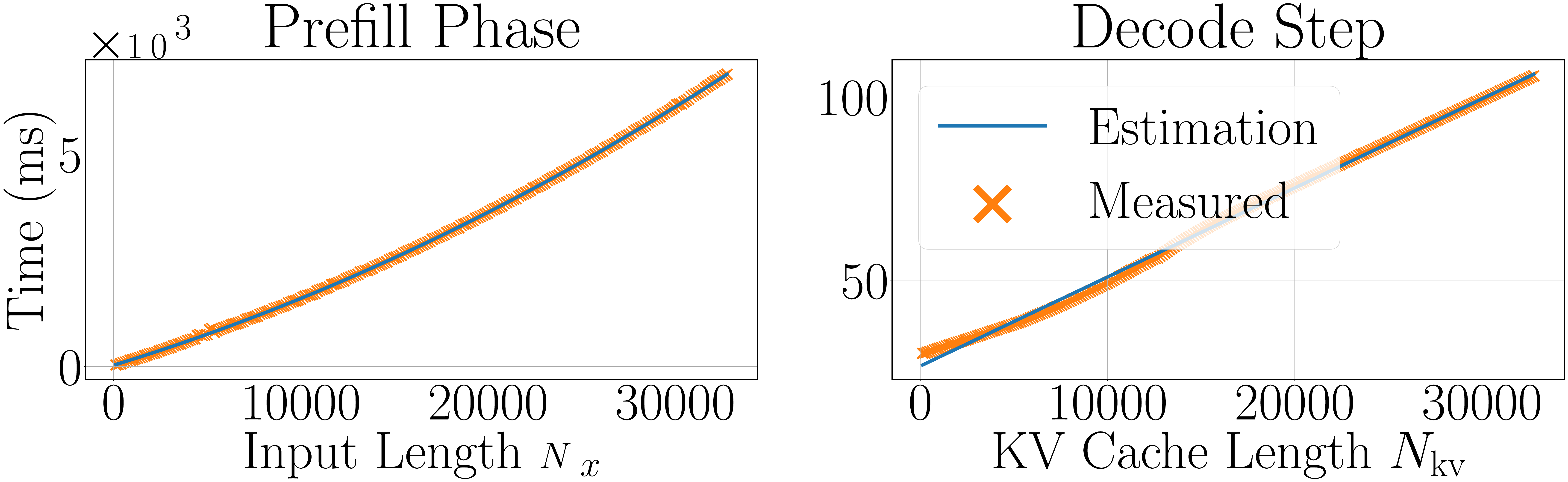}
    \caption{Fitted curves for estimating $\hat{t}_{\text{prefill-phase}},  \hat{t}_{\text{decoding-step}}$.}
    \label{fig:time_fitting}
\end{figure}

\begin{figure}[t]
    \centering
    \includegraphics[width=0.88\linewidth]{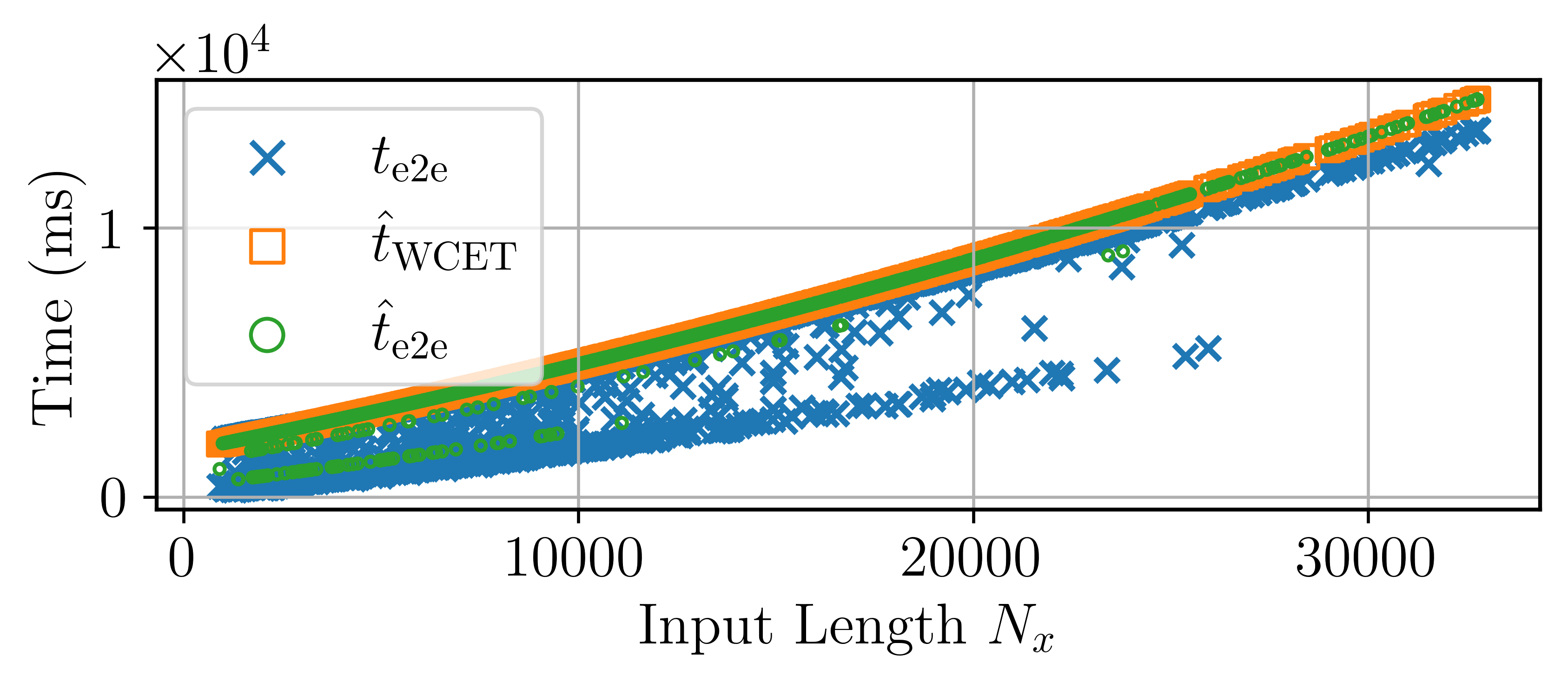}
    \caption{The performance of estimating $\hat{t}_{\text{e2e}}$ and $\hat{t}_{\text{WCET}}$.}
    \label{fig:e2e_time_estimation}
\end{figure}

We first evaluate the performance of estimating the $\hat{t}_{\text{prefill-phase}}$ and $\hat{t}_{\text{decoding-step}}^{i}$ using Mean Absolute Percentage Error (MAPE).
The MAPEs are $1.22\%$ and $1.69\%$ for the prefill phase and the decoding step, respectively, as shown in Fig.~\ref{fig:time_fitting}. 
Furthermore, we evaluate the performance of the end-to-end ETE, where $\texttt{Predict}(\cdot)$ with 512 buckets is utilized. The results of $\alpha=0$ and $N_{\text{max}}=64$ are presented in Fig.~\ref{fig:e2e_time_estimation}.
We can see that $\hat{t}_{\text{e2e}}$ is close to the actual $t_{\text{e2e}}$, while $\hat{t}_{\text{WCET}}$ effectively provides an upper bound of $t_{\text{e2e}}$, demonstrating the effectiveness of the proposed ETE.

\subsection{Comparison with Baselines}
\label{sec:compare_result}

\begin{figure}[t]
    \centering
    \includegraphics[width=0.88\linewidth]{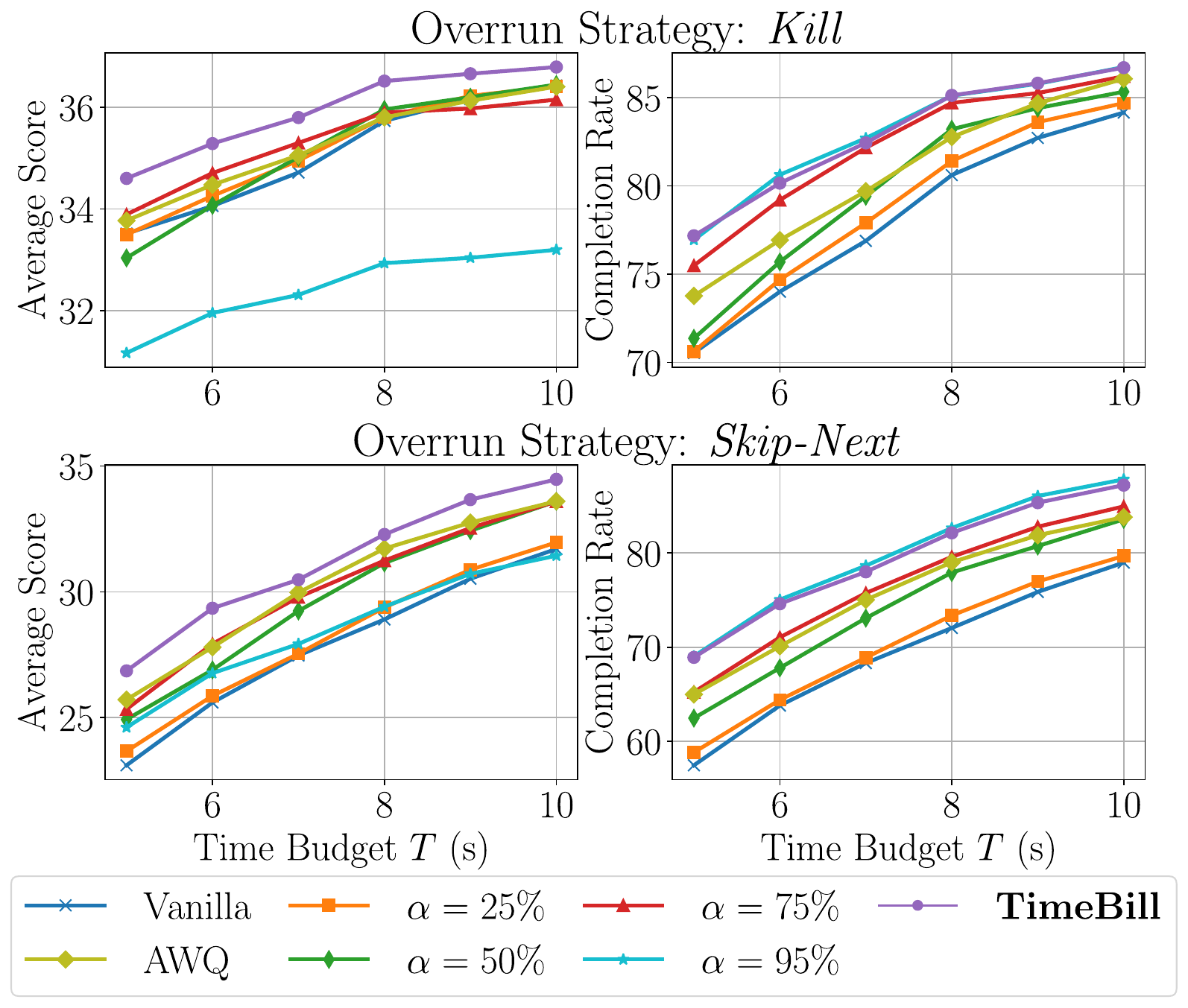}
    \caption{The average scores and completion rates of different approaches under \textit{Kill} and \textit{Skip-Next}.}
    \label{fig:comparison}
\end{figure}

We conduct experiments on six different time budgets, ranging from 5 s to 10 s in one-second increments. 
The average response performance scores and completion rates under the \textit{Kill} and \textit{Skip-Next} strategies are shown in Fig.~\ref{fig:comparison}. 
We can see that TimeBill achieves the state-of-the-art performance in average score and maintains a competitive task completion rate. 
Since the Vanilla often exceeds the time budget, it performs poorly with a low task completion rate and average score.
For KV cache eviction with fixed $\alpha$, as $\alpha$ increases, the task completion rates increase, while the average score first increases and then decreases. 
This is because when $\alpha$ is small, the benefit gained from allowing more inferences to finish within $T$ outweighs the loss in response performance, causing the average score to increase as $\alpha$ increases.
However, when $\alpha$ is large, it degrades the response performance significantly.
Furthermore, AWQ performs slightly better than the Vanilla. Moreover, TimeBill is orthogonal to it and can be effectively integrated with quantization.
In contrast, TimeBill balances the inference efficiency and response performance, thereby achieving the highest average response performance scores among tested approaches while attaining a similar task completion rate as $\alpha=95\%$. 

\subsection{Impacts of the Pessimistic Factor $k$}
\label{sec:exper_pess_factor}

\begin{figure}[t]
    \centering
    \includegraphics[width=0.85\linewidth]{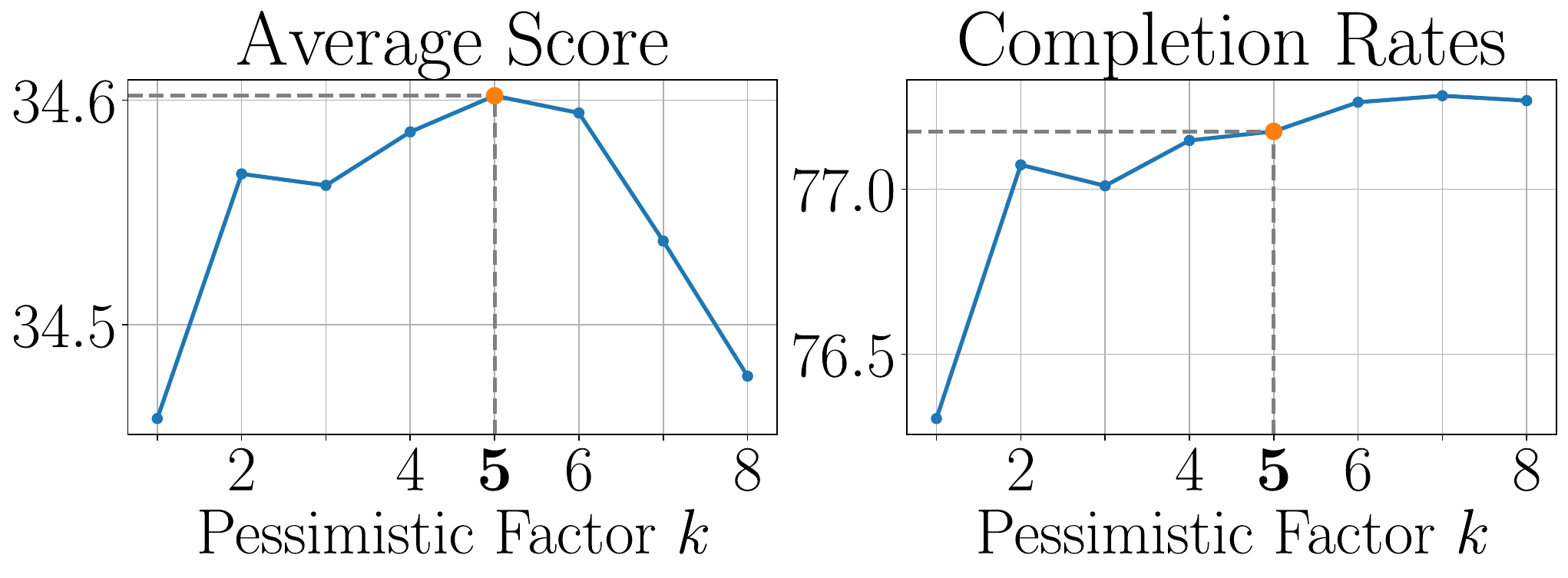}
    \caption{The average scores and completion rates with different pessimistic factors $k$ under the overrun strategy \textit{Kill}, where the time budget $T=5$ s.}
    \label{fig:impact_of_pessimistic}
\end{figure}

We conduct experiments to demonstrate the impact of $k$ under the \textit{Kill} strategy, where $k \in [1, 8]$, and the time budget $T=5$ s, as shown in Fig.~\ref{fig:impact_of_pessimistic}.
Since the more conservative $\hat{t}_{\text{WCET}}$ is, the higher the assurance of $t_{\text{e2e}} \leqslant \hat{t}_{\text{WCET}}$~\cite{Factor1}, which is consistent with the fact that a pessimistic factor $k = 5$ is common in hard real-time systems~\cite{Factor2}.
When $k$ is relatively small (1-5), increasing $k$ 
results in a higher $\alpha$ and allowing more inferences to be completed within $T$. 
Thus, both the average score and the completion rate increase. 
However, a large $k$ (6-8) leads to a large $\alpha$, causing significant degradation in response performance and leading to a decrease in the average score.
Therefore, $k$ should be carefully selected. 

\section{Conclusion}
\label{sec:conclusion}

In this paper, we propose TimeBill, a novel time-budgeted inference framework for LLMs. 
We present the problem formulation of time-budgeted inference for LLMs. 
We introduce a fine-grained RLP and a workload-guided ETE, enabling accurate end-to-end execution time prediction for LLMs. 
We develop a time-budgeted efficient inference method and provide the corresponding deployment.
Through extensive experiments, we demonstrate the state-of-the-art performance of TimeBill in improving both response performance and completion rate. 

\section*{Acknowledgments}

This work is supported in part by the NSF of China under Grants 62473254 and 62202287, and in part by the Open Research Fund of Peng Cheng Laboratory under Grant 2025KF1B0010.

\bibliography{aaai2026}

\end{document}